\documentclass[conference]{IEEEtran}
\IEEEoverridecommandlockouts
\usepackage{cite}
\usepackage{amsmath,amssymb,amsfonts}
\usepackage{algorithmic}
\usepackage{graphicx}
\usepackage{textcomp}
\usepackage{makecell}
\usepackage{multirow}
\usepackage{xcolor}
\def\BibTeX{{\rm B\kern-.05em{\sc i\kern-.025em b}\kern-.08em
    T\kern-.1667em\lower.7ex\hbox{E}\kern-.125emX}}
\begin{document}

\title{ Design and Analysis of Modular Pipe Climber-III with a Multi-Output Differential Mechanism
\thanks{All the authors are with Robotics Research Center, International Institute of Information Technology, Hyderabad-500032, India. (e-mail: vishnukmr001@gmail.com)}}

\author
{\IEEEauthorblockN{Vishnu Kumar\IEEEauthorrefmark{1}, Saharsh Agarwal\IEEEauthorrefmark{1}, Rama Vadapalli\IEEEauthorrefmark{1}, Nagamanikandan Govindan\IEEEauthorrefmark{1}, K Madhava Krishna\IEEEauthorrefmark{1}}
%
}
\maketitle

\begin{abstract}

This paper presents the design of an in-pipe climbing robot that operates using a novel `Three-output open differential'(3-OOD) mechanism to traverse complex networks of pipes. Conventional wheeled/tracked in-pipe climbing robots are prone to slip and drag while traversing in pipe bends. The 3-OOD mechanism helps in achieving the novel result of eliminating slip and drag in the robot tracks during motion. The proposed differential realizes the functional abilities of the traditional two-output differential, which is achieved the first time for a differential with three outputs. The 3-OOD mechanism mechanically modulates the track speeds of the robot based on the forces exerted on each track inside the pipe network, by eliminating the need for any active control. The simulation of the robot traversing in the pipe network in different orientations and in pipe-bends without slip shows the proposed design's effectiveness.


\end{abstract}



\section{Introduction}
Pipe networks are omnipresent, primarily used to  transport liquids and gases in industries and urban cities. Most often, the pipes are concealed to comply with the safety guidelines and to avoid risks. This makes inspection and maintenance of pipes very difficult. Buried pipes are highly prone to clogging, corrosion, scale formation, and crack initiation, resulting in leaks or damages that may lead to catastrophic incidents. 

Various In-Pipe Inspection Robots (IPIRs) \cite{b1} were proposed in the past to conduct regular preventive inspections to avoid accidents. Wall-pressed IPIRs with single, multiple and hybrid locomotion systems \cite{b2}, Pipe Inspection Gauges (PIGs) \cite{b3}, actively controlled IPIRs with articulated joints and differential drive units \cite{b4} were also extensively studied. Furthermore, bio-inspired robots with crawler, inchworm, walking mechanisms\cite{b5}, and screw-drive\cite{b6} mechanisms were also shown to be suitable for different requirements. However, most of them use active controlling methods to steer and manoeuvre inside the pipe. Dependence upon the robot's orientation inside the pipe added to the challenges, also leaving the robot vulnerable to slip if traction control methods are not involved. The Theseus \cite{b10}, PipeTron \cite{b12}, and PIRATE \cite{b13} robot series use separate segments for driving and driven modules that are interconnected by different linkage types. Each segments align or change the orientation for negotiating turns. Additionally, robust active steering makes such robots reliant on sensor data and heavy computation. 

\begin{figure}[ht!]
\centering
\includegraphics[width=3.3in]{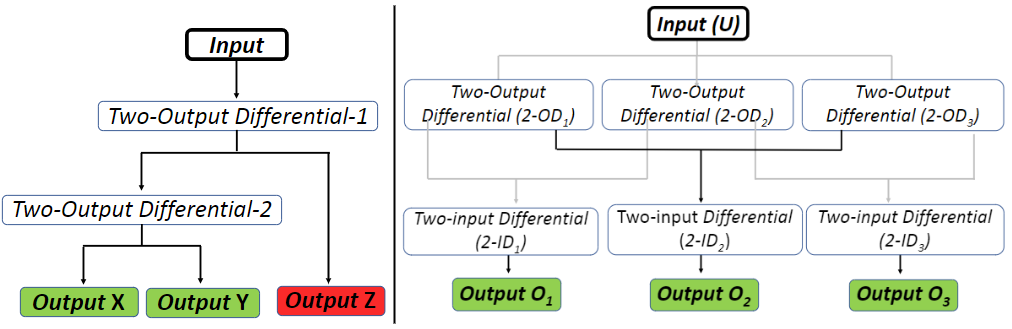}
\centering \space\space \space\space(a) \space\space \space\space\space \space\space\space\space\space\space\space\space\space\space \space\space \space \space\space \space\space \space\space \space\space \space\space(b)
\caption{\footnotesize (a) Layout of the previous Three-output differentials(3-OD's) (b) Layout for Three-Output Open Differential (3-OOD)}
\label{Schematics}
\vspace{-0.1in}
\end{figure}

Pipe climbing robots with three symmetrical modules are more stable and provide better mobility. Our earlier proposed modular pipe climbers \cite{b14,b15} have used three driving tracks arranged symmetrically to each other, similar to MRINSPECT series of robots \cite{b16,b17,b18}. In such robots, to actively control the three tracks, their velocities were pre-defined for the pipe-bends. This posed a limitation for the robot to negotiate pipe-bends only at a particular orientation corresponding to the pre-defined velocities\cite{b14,b15,b16,b17,b18}. In real applications, the robot's orientation change if it experiences slip in the tracks during motion. This limitation can be solved by using a passively operated differential mechanism to control the robot. MRINSPECT-VI \cite{b19,b20} uses a multi-axial differential gear mechanism to control the speeds of the three modules. However, for the division of the driving torque and speed to the three modules, the layout of the differential shown in Fig. \ref{Schematics}(a) is used. This strategy made the first output (Z) to rotate faster than the other two outputs (X and Y), making output Z easily affected by slip \cite{b19}. This is caused because the outputs of the differential does not share equivalent kinetics with the input. Other previously proposed solutions for Three-output differentials (3-OD's) \cite{b21,b22} also followed a similar differential layout, as shown in Fig. \ref{Schematics}(a).
 \vspace{-0.2in}

Our `Three-Output Open Differential' eliminates the mentioned limitation by realizing equivalent output to input kinetic relations \cite{b23}. In the devised layout all the three outputs are equally affected by the input, shown in Fig. \ref{Schematics}(b). This contributes for the robot to eliminate slip and drag in any orientation of the robot during its motion. Additionally, the differential mechanism in the pipe climber enhances the ease of use by reducing the dependency on the active controls to manoeuvre through the pipe networks.


\section{Design of the Modular Pipe Climber-III}

\subsection{Structure of the Robot}

\begin{figure}[ht!]
\centering
\includegraphics[width=3in]{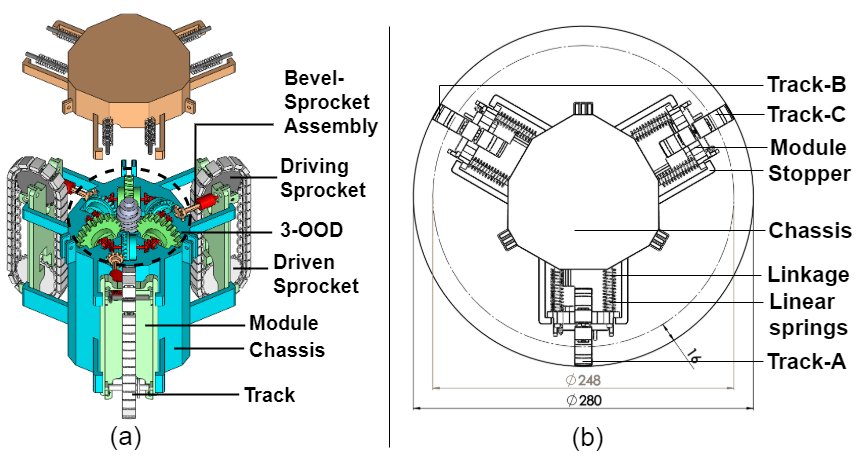}
\vspace{-0.15in}
\caption{\footnotesize (a) Robot structure (b) Top view of the robot }
\label{1}
\vspace{-0.15in}
\end{figure}

The CAD model of the proposed robot, Modular Pipe Climber-III, is shown in Fig.~\ref{1}(a). The differential mechanism (3-OOD) is positioned inside the robot's central chassis and it drives each track individually through their respective driving sprockets. The output velocity of the 3-OOD mechanism is translated to the driving sprockets through a bevel gear setup. The detailed view of the 3-OOD mechanism is shown in Fig.\ref{2} (a). An irregular nonagon central chassis of the robot houses three modules (A, B, C) circumferentially 120$^\circ$ apart from each other, illustrated in Fig. \ref{1}(b). Each module houses a track to encompass the driving and driven sprockets, and a spring-loaded mechanism to attach with the chassis. The linear springs supported with a guiding rod (Linkage) between the modules and the chassis push the modules radially outwards when they are compressed. The stoppers restrict the radial extension of the modules within the permissible limits. When the robot is deployed inside the pipe, the spring-loaded tracks undergo deflection and press against the pipe's inner walls, providing them the necessary traction to move or hold position. Each module in the robot can also compress asymmetrically, as detailed in section III.


\vspace{-0.15in}


\subsection{Three-Output Open Differential (3-ODD)}

\begin{figure}[ht!]
\centering
\includegraphics[width=3in]{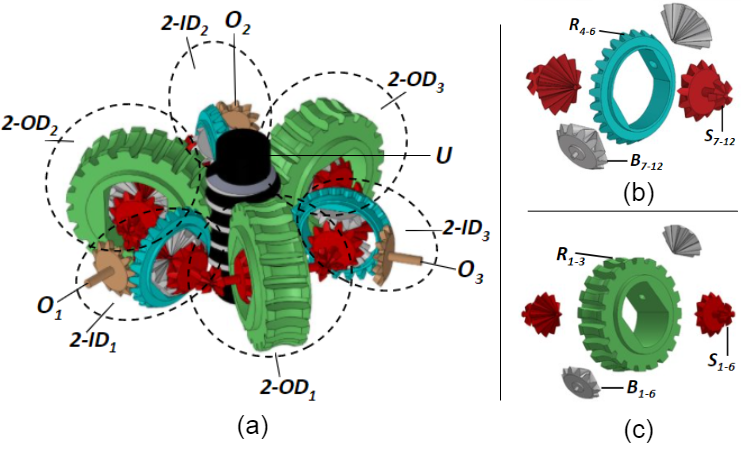}
\vspace{-0.15in}
\caption{\footnotesize (a) 3-OOD mechanism (Isometric view) (b) Two-output differential $(2-OD_{1-3})$ (c) Two-input differential $(2-ID_{1-3})$}
\label{2}
\vspace{-0.05in}
\end{figure}

The Three-Output Open Differential (3-OOD) is the principal constituent of the proposed robot, Fig. \ref{2}(a). The 3-OOD mechanism comprises a single input $(U)$, three two-output open differentials $(2-OD_{1-3})$, three two-input open differentials $(2-ID_{1-3})$ and three outputs $(O_{1-3})$. The differential’s input $(U)$ is located at the central axis of the robot chassis. The three $2-OD_{1-3}$, shown in Fig. \ref{2}(c) are arranged symmetrically around the input $(U)$, with an angle of $120{^\circ}$ between any two. The $2-ID_{1-3}$, shown in Fig. \ref{2}(b) are fitted radially in-between $2-OD_{1-3}$. The single output of $2-ID_{1-3}$ form the three outputs $(O_{1-3})$ of the 3-OOD. $2-OD_{1-3}$ comprises of gear elements such as ring gears $(R_{1-3})$, bevel gears $(B_{1-6})$ and side gears $(S_{1-6})$, while $2-ID_{1-3}$ include ring gears $(R_{4-6})$, bevel gears $(B_{7-12})$ and side gears $(S_{7-12})$. The side gears $(S_{1-6})$ of $2-OD_{1-3}$ is meshed with their adjacent side gears $(S_{7-12})$ of $2-ID_{1-3}$, to transfer the torque and speed from $2-OD_{1-3}$ to it's adjacent $2-ID_{1-3}$. The input $(U)$ of the worm gear provide motion to $2-OD_{1-3}$ simuntaneously. Each two-output differential $(2-OD_{1-3})$ then transfer received motion to its adjacent two-input differentials $(2-ID_{1-3})$, depending on the load conditions experienced by its respective side gears $(S_{1-6})$. The motion received by the side gears $(S_{7-12})$ of $2-ID_{1-3}$ further translates them to the three outputs ($O_{1-3})$. The six differentials ($2-OD_{1-3}$ and $2-ID_{1-3})$ work together to translate motion from the input to the three outputs ($O_{1-3})$.
 
\vspace{-0.15in}
 
When  $O_{1-3}$ experience different loads, the side gears ($S_{7-12})$ in $2-ID_{1-3}$ transfers different loads to the side gears $(S_{1-6})$ of $2-OD_{1-3}$. Under this condition, $2-OD_{1-3}$ translates differential speed to its adjacent $2-ID_{1-3}$. When $O_{1-3}$ experience equal load or no load, all the side gears experience the same load making them rotate at the same speed and torque. 

As outlined in Fig. \ref{Schematics}(b), all the outputs ($O_1$, $O_2$, $O_3$) share equal kinetics with the input. In addition, the outputs also share identical kinetics with each other. This results in the change of loads in one of the outputs imposing an equal effect on the other two outputs that are undisturbed. The outputs operate with equal speeds when there is no load or equal load acting on all the outputs. The 3-OOD mechanism operates its outputs with differential speed when the outputs are under varied loads. When one of the outputs is operating at a different speed while the other two outputs are experiencing the same load, then the two outputs with the same loads will operate with equal speeds.
 
The 3-OOD mechanism designed for the Modular Pipe Climber III, advances its the three tracks with equivalent speeds when moving inside a straight pipe. But while manoeuvring inside the pipe-bends, the differential modulates the track speed of the robot such that the track travelling the longer distance rotates faster than the track travelling the shorter distance.
\vspace{-0.02in}
\section{Kinematics and Dynamics}
\vspace{-0.01in}
\subsection{Kinematics and Dynamics of the 3-OOD mechanism}
\vspace{-0.02in}
\begin{figure*}[ht!]
\centering
\includegraphics[width=2in]{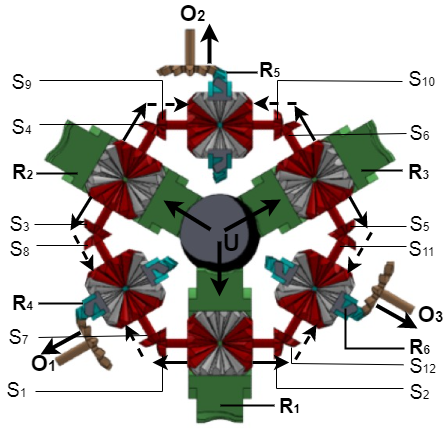}
\includegraphics[width=2in]{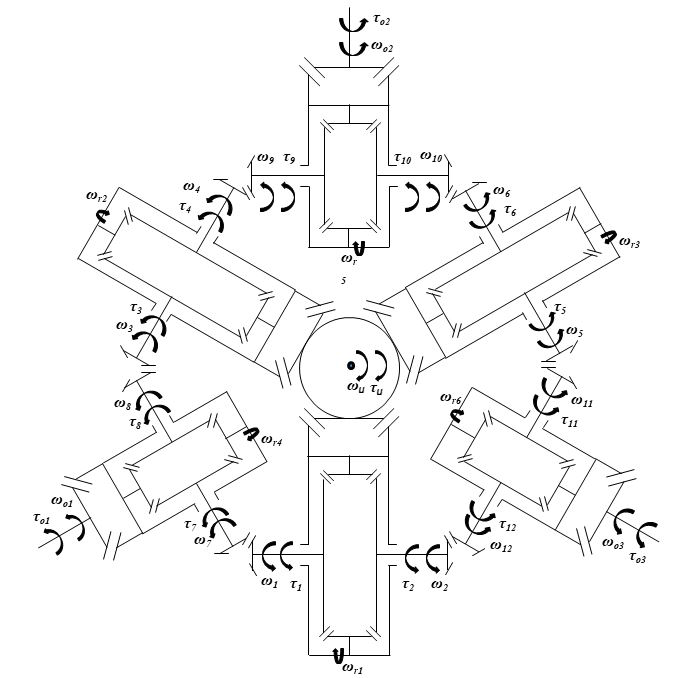}
\includegraphics[width=2.4in]{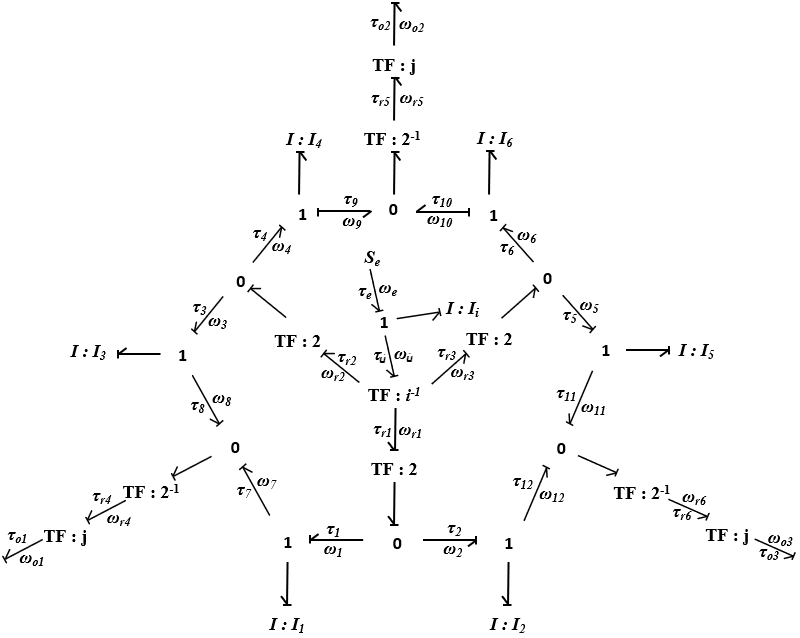}
\vspace{0.01in (a)\space\space\space\space\space\space\space\space\space\space\space\space\space\space\space\space\space\space\space\space\space\space\space\space\space\space\space\space\space\space\space\space\space\space\space\space(b) \space\space\space\space\space\space\space\space\space\space\space\space\space\space\space\space\space\space\space\space\space\space\space\space\space\space\space\space\space\space\space\space\space\space\space\space\space\space\space\space\space\space\space\space\space(c)}
\vspace{-0.08in}
\caption{\footnotesize Three-Output Open Differential: (a) Power flow diagram (b) Kinematic scheme (c) Bond  Graph  Model}
\label{Kinematic}
\vspace{-0.13in}
\end{figure*}

Fig.~\ref{Kinematic}(a) represents an illustration of the direction of speed and torque flow in the differential. The kinematic scheme shown in Fig.~\ref{Kinematic}(b) illustrates the connectivity of the links and joints of the 3-OOD mechanism. The Three-Output Open Differential's kinematic and dynamic equations are formulated using the bond graph model shown in Fig.~\ref{Kinematic}(c). 

The input angular velocity ($\omega_u$) from the motor is equally distributed to the three ring gears of the two-output differentials as $\omega_{r1}$, $\omega_{r2}$ and $\omega_{r3}$. They rotate at equal angular velocities and with equal torque i.e. $1/k$ times $\omega_u$ and $k/3$ times $\tau_u$, where $1/k$ is the gear ratio of the input to the ring gears ($1/k$ = $1/20$). Moreover, a two-output differential dictates that the angular velocity of its ring gear is always the average of the angular velocities of its two side gears. These two side gears can rotate at different speeds while maintaining equal torque \cite{b24}.

\vspace{-0.15in}
\begin{equation}
\omega _u=\frac{k(\omega _1+\omega _2)}{2}=\frac{k(\omega _3+\omega _4)}{2}=\frac{k(\omega _5+\omega _6)}{2},
\label{6velocities_S1-6-I}
\end{equation}
\vspace{-0.18in}


The angular velocities $\omega_{r1}$, $\omega_{r2}$ and $\omega_{r3}$ are then translated to their side gears. Side gears $(S_1$, $S_7$), $(S_3$, $S_8$) and $(S_5$, $S_{11}$) are connected such that they do not have any relative motion between the gears of the same pair, i.e., angular velocities of side gears of the same pair is always equal $(\omega_1$$=$$\omega_7)$, $(\omega_3$$=$$\omega_8)$ and $(\omega_5$$=$$\omega_{11})$. Substituting in (\ref{6velocities_S1-6-I})

\vspace{-0.20in}
\begin{equation}
\begin{split}
\omega _7\hspace{-0.04in}=\hspace{-0.04in}\omega _1\hspace{-0.04in}=\hspace{-0.04in}\frac{2(\omega _i)}{k}\hspace{-0.04in}-\hspace{-0.04in}\omega _2,\hspace{0.03in}\omega _8\hspace{-0.04in}=\hspace{-0.04in}\omega _3\hspace{-0.04in}=\hspace{-0.04in}\frac{2(\omega _i)}{k}\hspace{-0.04in}-\hspace{-0.04in}\omega _4,\hspace{0.03in}&\omega _{11}\hspace{-0.04in}=\hspace{-0.04in}\omega _5\hspace{-0.04in}=\hspace{-0.04in}\frac{2(\omega _i)}{k}\hspace{-0.04in}-\hspace{-0.04in}\omega _6
\end{split}
\label{12velocity_S1-12-I}
\end{equation}
\vspace{-0.17in}

Similarly, the output angular velocities $\omega_{O1}$, $\omega_{O2}$, $\omega_{O3}$ are derived from the velocities $\omega_{r4}$, $\omega_{r5}$, $\omega_{r6}$ of the respective ring gears $R_4$, $R_5$, and $R_6$. The output to side gear relation is equated with (\ref{12velocity_S1-12-I}), to attain the angular velocity equation for the input to the outputs.

{\vspace{-0.2in}
\begin{equation}
\begin{split}
& \hspace{-0.06in} \omega _{O1} \hspace{-0.04in}=\hspace{-0.04in} \frac{2j(\omega_u)}{k} \hspace{-0.04in}-\hspace{-0.04in} \frac{j(\omega _2+\omega _4)}{2},\hspace{0.025in}
\omega _{O2}\hspace{-0.04in}=\hspace{-0.04in} \frac{2j(\omega_u)}{k} \hspace{-0.04in}-\hspace{-0.04in} \frac{j(\omega _3+\omega _5)}{2},\\ &
\hspace{0.7in} \omega _{O3} \hspace{-0.04in}=\hspace{-0.04in} \frac{2j(\omega _u)}{k} \hspace{-0.04in}-\hspace{-0.04in} \frac{j(\omega _1+\omega _6)}{2}
\end{split}
\label{17velocity_O3-I}
\end{equation}
}
where $\omega_u$ is the angular velocity of the input $(U)$, $j$ is the gear ratio of the ring gears $(R_{4-6})$ to the outputs ($j$ = 2:1), while $\omega_{O1}$,  $\omega_{O2}$ and $\omega_{O3}$ are angular velocities of the output. $\omega_1$, $\omega_2$, $\omega_3$, $\omega_4$, $\omega_5$ and $\omega_6$ are the respective angular velocities of the side gears $(S_1$, $S_2$, $S_3$, $S_4$, $S_5$, $S_6)$. Therefore, the output angular velocities ($\omega_{O1}$, $\omega_{O2}$, $\omega_{O3}$) are dependent on the input angular velocity ($\omega_u$) and the side gear outputs ($\omega_2$, $\omega_4$), ($\omega_3$, $\omega_5$) and ($\omega_1$, $\omega_6$).  

Meanwhile, the torques $\tau_{R4}$, $\tau_{R5}$ and $\tau_{R6}$ of the ring gears $(R_4$, $R_5$ and $R_6)$ is the sum of the torques of their corresponding side gears $S_7$, $S_8$, $S_9$, $S_{10}$, $S_{11}$ and $S_{12}$. Similar to angular velocity relation for input to outputs, by equating the ring gears to side gears relation with the output to the ring gear relation, a relationship between output torques ($\tau_{O1}$, $\tau_{O2}$, $\tau_{O3}$) to the input torque ($\tau_{u}$) is obtained.
\vspace{-0.05in}
\begin{equation}
\begin{split}
& \tau_{O1}=\frac{k(\tau_u)}{3j}-\frac{(I_1\Dot{\omega}_7+I_3\Dot{\omega}_8)}{j}, \ \hspace{0.2in}  \tau_{O2}=\frac{k(\tau_u)}{3j}\hspace{0.025in}- \\ &\frac{(I_4\Dot{\omega}_9+I_6\Dot{\omega}_{10})}{j}, \ \hspace{0.2in}  \tau_{O3}=\frac{k(\tau_u)}{3j}-\frac{(I_2\Dot{\omega} _{12}+I_5\Dot{\omega}_{11})}{j}
\label{23torque_O3-I}
\end{split}
\end{equation}

where $I_1$,$I_2$,$I_3$,$I_4$,$I_5$,$I_6$ are the inertia exhibited by the side gears, $\Dot{\omega}_7$, $\Dot{\omega}_8$, $\Dot{\omega}_9$, $\Dot{\omega}_{10}$, $\Dot{\omega}_{11}$, $\Dot{\omega}_{12}$ are the angular acceleration of the respective side gears and $\tau_u$ is the input torque.

The 3-OOD mechanism has three degrees of freedom in its output. Equations (\ref{17velocity_O3-I}) and (\ref{23torque_O3-I}) validates that the behavior of each output is impacted by the input ($U$) as well as the other two outputs.

\textbf{\small Equal speeds and torque:\normalsize} Identical side gears $(S_{7-12})$  exhibit equal inertia $(I_1=I_2=I_3=I_4=I_5=I_6)$. The side gears $(S_{7-12})$ operate with equal angular velocity $(\omega _n= \frac{\omega _i}{k}$, where $n$ ranges from $1$ to $12)$ and angular acceleration when all three outputs $(O_{1-3})$ experience equal loads. Substituting these relations in \eqref{17velocity_O3-I} and \eqref{23torque_O3-I} we get,
\vspace{-0.1in}
\begin{equation}
{\bf\omega _{O1}}={\bf\omega _{O2}}={\bf\omega _{O3}}=\frac{j(\omega _u)}{k}
\label{27Velocity_O1=O2=O3}
\end{equation}
\begin{equation}
{\bf\tau _{O1}}={\bf\tau _{O2}}={\bf\tau _{O3}}=\frac{k(\tau_u)}{3j} - \frac{2(I_1\dot{\omega}_1)}{j}
\label{28Torque_O1=O2=O3}
\end{equation}

Equations \eqref{27Velocity_O1=O2=O3} and \eqref{28Torque_O1=O2=O3} presents the novelty of the differential to provide equal motion characteristics in all three outputs that experiences equal loads or when left unconstrained. But when the gear components experience a resistance across a junction, the angular velocity and the torque changes depending on the external resistive force.

\subsection{Track Velocity}

\begin{table}[t]
{
\caption{\footnotesize Theoretical velocities of $v_A$,$v_B$ and $v_C$ in Elbow and U-section for various robot orientations ($\mu$) }}
\vspace{-0.2in}
\begin{center}
{
\label{tbl:myLboro}
}
  \centering
  \scalebox{0.645}{
  \begin{tabular}{ | m{2.6cm} | m{2.6cm} | m{2.6cm} |  m{2.6cm} | }
    \hline
    \centering
   Robot Orientation &  \begin{minipage}{2.5cm}
    \centering
    \vspace{0.03in}
      \includegraphics[width=15mm]{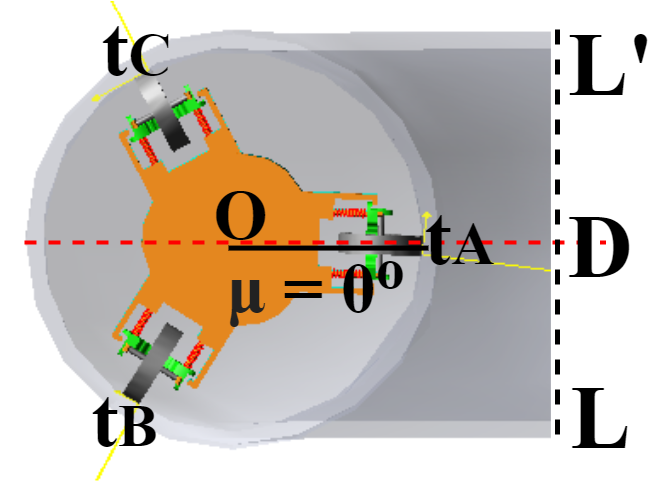}
      {\\ \footnotesize $\alpha$ = $0^\circ$}
    \end{minipage} & 
    \begin{minipage}{2.5cm}
    \centering
    \vspace{0.03in}
      \includegraphics[width=15mm]{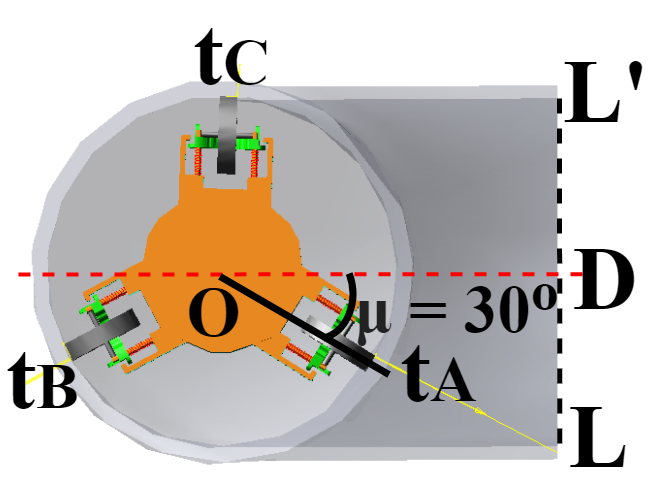}
     {\\ \footnotesize $\alpha$ = $30^\circ$}
    \end{minipage}  &   
    \begin{minipage}{2.5cm}
    \centering
    \vspace{0.03in}
      \includegraphics[width=15mm]{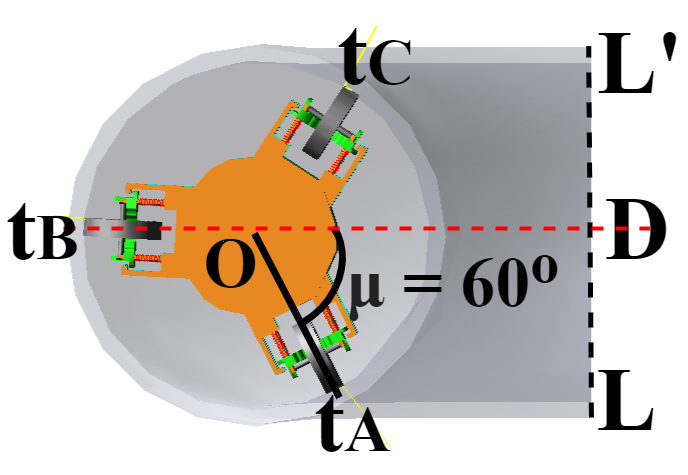}
      {\\ \footnotesize $\alpha$ = $60^\circ$}
    \end{minipage} \\ \hline
    \centering
   Velocity ratio's ($v_A$:$v_B$:$v_C$)
    &
    \makecell{
     0.670 : 1.165 : 1.165}
    & 
      \makecell{
       0.715 : 1.285 : 1}
    &
    \makecell{
      \\ 0.835: 1.329: 0.835}
  
    \\ \hline
       Inner-module velocity (in mm/sec)
    &
    \makecell{
      $v_{tA}$ = 33.69
    }
    & 
    \makecell{
     $v_{tA}$ = 35.90,
    \\ $v_{tC}$ = 50.24 \tiny{(Center module})}  
    &
    \makecell{
     $v_{tA}$ = $v_{tC}$ = 41.96}
 
    \\ \hline  
     Outer-module velocity (in mm/sec)
    &
    \makecell{
     $v_{tB}$ = $v_{tC}$ = 58.51}
    & 
    \makecell{
     $v_{tB}$ = 64.57}
    &
    \makecell{
     \\ $v_{tB}$ = 66.78}
    \\ \hline
    \end{tabular}}
    \vspace{-0.1in}
\end{center}
\vspace{-0.1in}
\end{table}

\begin{figure}[htbp]
\centerline{\includegraphics[scale=0.21]{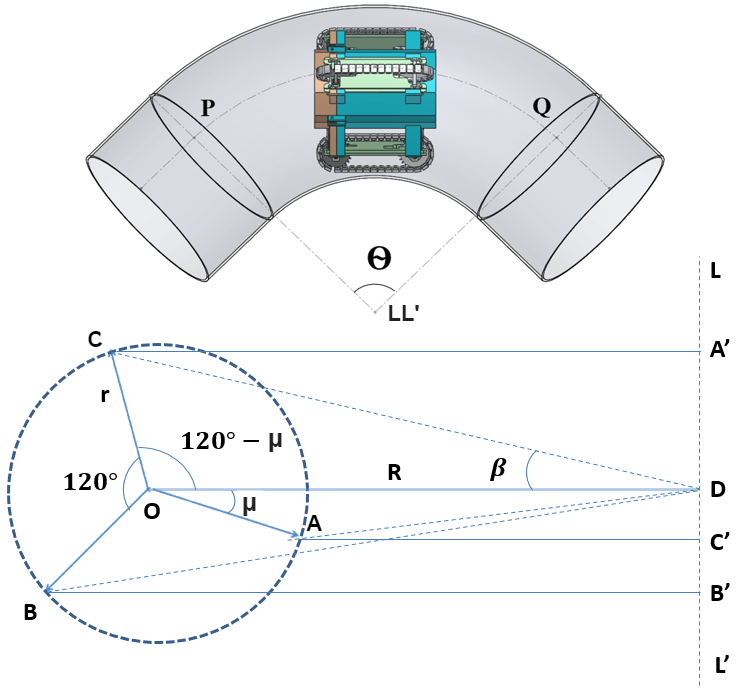}}
\vspace{-0.12in}
\caption{Two-dimensional representation of robot's orientation in pipe bends}
\label{Figure9}
\vspace{-0.2in}
\end{figure}

 The output velocity of the differential outputs are equal to the speed of the driving module sprockets.Therefore, the output velocities ${\omega_{O1}}$, ${\omega_{O2}}$ and ${\omega_{O3}}$ of the differential are translated into track speeds $v_{tA}$, $v_{tB}$ and $v_{tC}$. The input speed for the robot is 120 rpm, thereby translating 12 rpm to the outputs under equal loading conditions. The sprocket diameter is constant ($D_s$= 80 mm) for all the three tracks.

When the robot travels in straight pipes, the tracks experience equal loads in all three tracks i.e., $\omega_{t1}$=$\omega_{t2}$=$\omega_{t3}$
\vspace{-0.04in}
\begin{equation}
v_{tA}= v_{tB}=v_{tC} = \frac{\pi\times{D_s}\times12}{60} = 50.24 \textrm{mm/sec} 
\label{eqnxx}
\end{equation}
The average speed of the three tracks are always equal to the speed of the robot, $v_{R}$. 
{
\vspace{-0.05in}
\begin{equation}
v_R = \frac{v_{tA}+v_{tB}+v_{tC}}{3} = 50.24 \textrm{mm/sec} 
\label{eqn}
\end{equation}
}
The speed of the tracks required to negotiate a pipe-bend depends on the orientation of the robot with respect to the bend direction. Ho Moon Kim et al. \cite{b20}, in their paper suggested a method for calculating the respective velocities of the three tracks inside pipe bends. Assuming that the robot enters the pipeline with the configuration shown in Fig.~\ref{Figure9}, {\small$\Theta$} is the angle of the pipe bend, R is the radius of curvature of the pipe bend and r is the radius of the pipe. The velocity of the track A is derived from Fig.~\ref{Figure9}
\vspace{-0.05in}
\begin{equation}
v_{tA} = v_R(\frac{R_A-r\cos({\mu})}{R_A})
\label{eqn01}
\vspace{-0.05in}
\end{equation}

In a similar way, the velocities $v_{tB}$ and $v_{tC}$ for their respective tracks B and C are obtained. The robot is inserted at different orientation of the modules with respect to OD. The modulated speeds for the tracks are calculated for bend-pipes in orientations $\mu$ = $0^\circ$, $\mu$ = $30^\circ$, $\mu$ = $60^\circ$, shown in table I.

\subsection{Asymmetrical Compression}

\begin{figure}[ht!]
\vspace{-0.1in}
\centering
\includegraphics[width=3in]{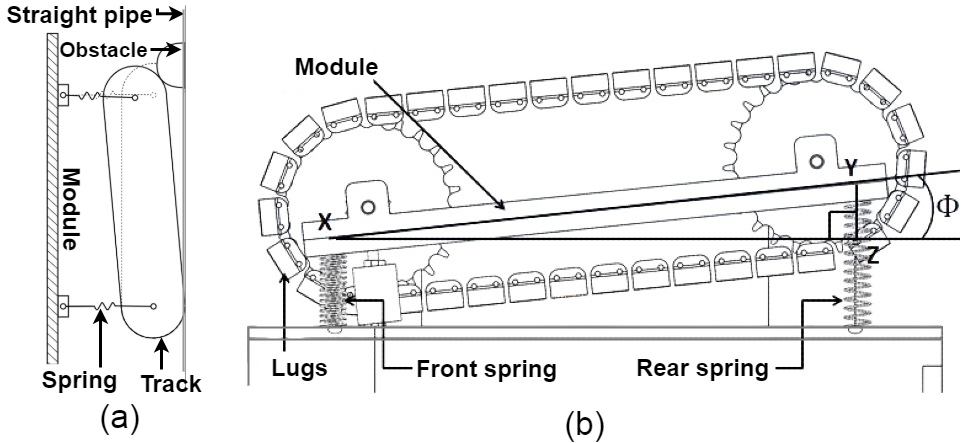}
\vspace{-0.12in}
\caption{\footnotesize (a) Robot overcoming obstacle (b) Asymmetrical module compression}
\label{asym(1)}
\vspace{-0.1in}
\end{figure}

The linear springs in the module provides robot the flexibility to negotiate bends easily. The maximum compression possible in each module is $16 mm$. There are additional tolerances in the module holes, so that asymmetrical compression is possible. This helps the robot to overcome obstacles and irregularities in the pipe network that it may face in real world applications, shown in Fig.~\ref{asym(1)}(a). In Fig.~\ref{asym(1)}(b), the front end of the module is compressed completely whereas the rear end is in its maximum extended state possible inside the pipe. The maximum asymmetrical compression allowed in a single module of the robot is shown in Fig.~\ref{asym(1)}(b). Thus, $\phi$ is the maximum angle the module can compress asymmetrically.

In $\bigtriangleup$$XYZ$, $YZ$ is perpendicular to $XZ$ and $\angle$ $YXZ$ is $\phi$, Thus, the maximum permissible asymmetrical bend is

\vspace{-0.1in}
\begin{equation}
\phi = \tan^{-1}(\frac{YZ}{XZ}) = \tan^{-1}(\frac{12}{150}) = 4.574^\circ
\label{eqn4}
\end{equation}

\vspace{-0.1in}



\section{Dynamic Simulations of the Pipe Climber }

Simulations were conducted to analyze and validate the motion capabilities of the robot in various pipe networks. The same will give us more insights into the dynamics and behaviour of the developed Modular Pipe Climber-III in real testing environments. Hence, multi-body dynamic simulations was performed in MSC Adams by converting the design parameters into a simplified simulation model. To reduce the number of moving components in the model and to decrease the computational load, the tracks were simplified into roller wheels. Each module houses three roller wheels in the simplified model. Therefore, the contact patch provided by the tracks to the pipe walls are reduced from 10 contact regions to 3 contact regions per module. The simulation parameters such as the track velocities ($v_{tA}$, $v_{tB}$, and $v_{tC}$) and the module compression for each track A,B and C were analyzed. Simulations were performed by inserting the robot in three different orientations of the module ($\mu$ = 0${^\circ}$,  $\mu$ = 30${^\circ}$, $\mu$ = 60${^\circ}$) in both the straight pipes and pipe-bends.
\begin{figure}[htbp]
\vspace{-0.08in}
\centerline{\includegraphics[scale=0.22]{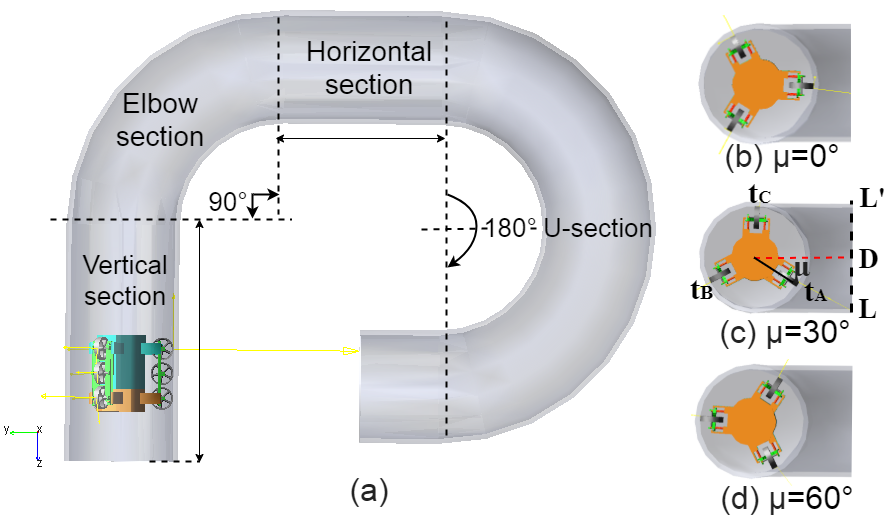}}
\vspace{-0.1in}
\caption{(a) Pipe network; Robot orientations: (b) $\mu$=0$^\circ$, (c) $\mu$=30$^\circ$, (d) $\mu$=60$^\circ$}
\label{Pipe}
\vspace{-0.05in}
\end{figure}
The robot is simulated inside a pipe-network designed according to ASME B16.9 standard NPS 11 and schedule 40. The simulations were conducted for four test case scenarios in the pipe network consisting of Vertical section, Elbow section (90${^\circ}$ bend), Horizontal section and the U-section (180${^\circ}$) for different orientations ($\mu$ = 0${^\circ}$,  $\mu$ = 30${^\circ}$, $\mu$ = 60${^\circ}$) of the robot, refer Fig.~\ref{Pipe}. The total distance of the pipe structure is $D_{pipe}$ = 3,023.49mm. The distance travelled by the robot ($D_{R}$) in pipe is calculated from center of the robot body and the track's individual distance travelled is calculated from the center roller wheel mounted in each modules. Therefore, we get the total robot's path, by subtracting the robot's length from $D_{pipe}$ (i.e., $D_{R}$ =$D_{pipe}$ - $(L_R)$ = 2,823.49mm, where $L_R$= 200mm).The robot's path in vertical climbing and the last horizontal section is measured by subtracting $L_R/2$ from their respective section length. The input ($U$) of the 3-OOD is given a constant angular velocity of 120rpm ($\omega_u$= 120rpm) and motion of the robot including the track velocities ($v_{tA}$, $v_{tB}$, and $v_{tC}$) are studied in the simulation.
\vspace{-0.1in}
\begin{figure}[htbp]
\centerline{\includegraphics[scale=0.18]{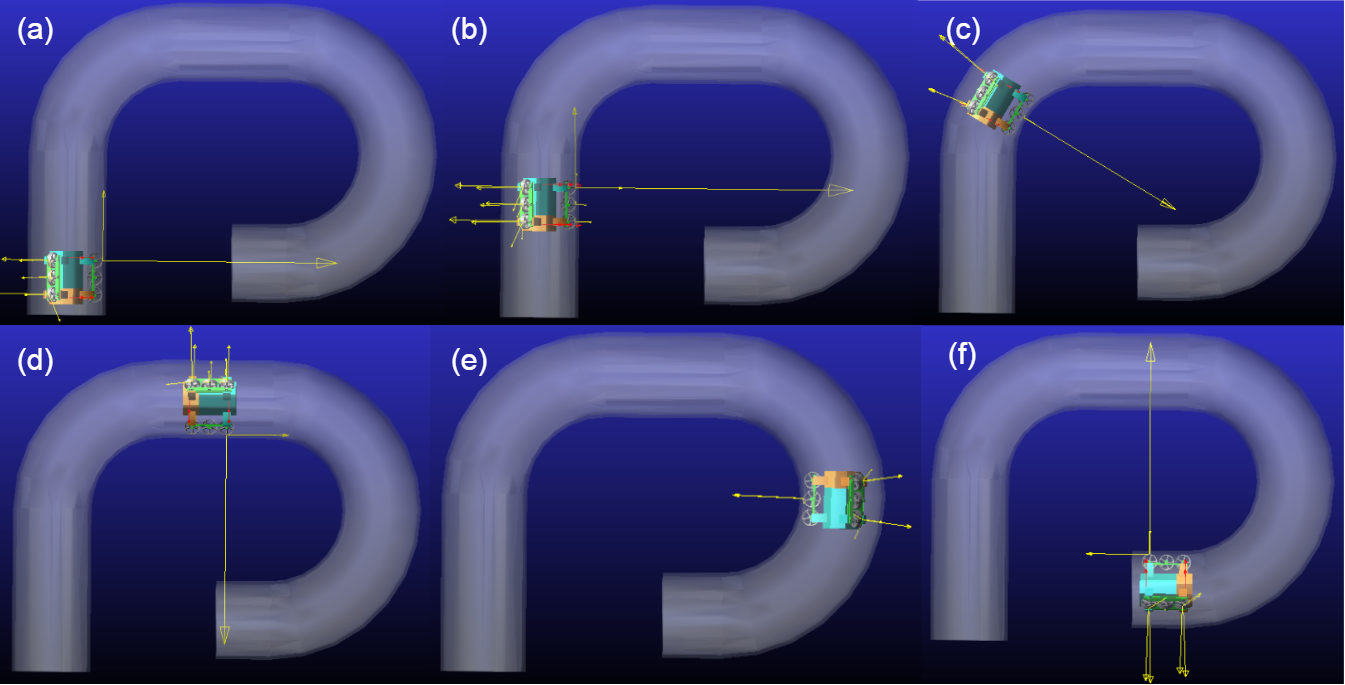}}
\vspace{-0.05in}
\caption{Simulation of the robot}
\label{Rsim}
\vspace{-0.15in}
\end{figure}

\subsection{Vertical section and Horizontal section}

In the vertical section and horizontal section, the robot follows a straight path. Therefore, the tracks experience equal loads on all three modules in both the test cases. As a result, the differential provides equal velocities for all the three tracks, equivalent to the robot's average velocity $v_R$. The observed track velocities in the simulation for the orientation $\mu$ = 0$^\circ$, is $v_R$= $v_{tA}$= $v_{tB}$= $v_{tC}$= 50.03 mm/sec. Similarly, for $\mu$ = 30$^\circ$, the velocities are $v_R$= $v_{tA}$= $v_{tB}$= $v_{tC}$= 50.22 mm/sec and for $\mu$ = 60$^\circ$, the velocities are $v_R$= $v_{tA}$= $v_{tB}$= $v_{tC}$= 51.36 mm/sec. Therefore, all the values correspond to the theoretical results shown in \eqref{eqn}, with an absolute percentage error (APE) lesser than 2.2\%. This error quantifies the actual amount of deviation from the theoretical value \cite{b25}. 
\vspace{-0.01in}
\begin{equation}
\small {APE = (\frac{Simulation \; velocity-Theoretical \; velocity}{Theoretical \; velocity})\times 100\% } 
\label{eqn02}
\end{equation}

To negotiate an initial length of 550 mm ($D_{R}$ = 550 - $(L_R/2)$= 450 mm) in vertical climbing, the robot takes 0 to 9 seconds. Figures~\ref{Rsim}(a) and~\ref{Rsim}(b) represents the robot's ability to climb the pipe vertically against gravity. Starting from 24 seconds till 31 seconds, the robot moves a distance of 350 mm in the first horizontal section, shown in Fig.~\ref{Rsim}(d). In the smaller horizontal section of distance 150mm ($D_{R}$ = 150 - $(L_R/2)$= 50 mm), the robot moves from 59 to 60 seconds, shown in Fig.~\ref{Rsim}(f).

\vspace{0.15in}
\begin{figure}[ht!]
\centering
\includegraphics[width=3.4in]{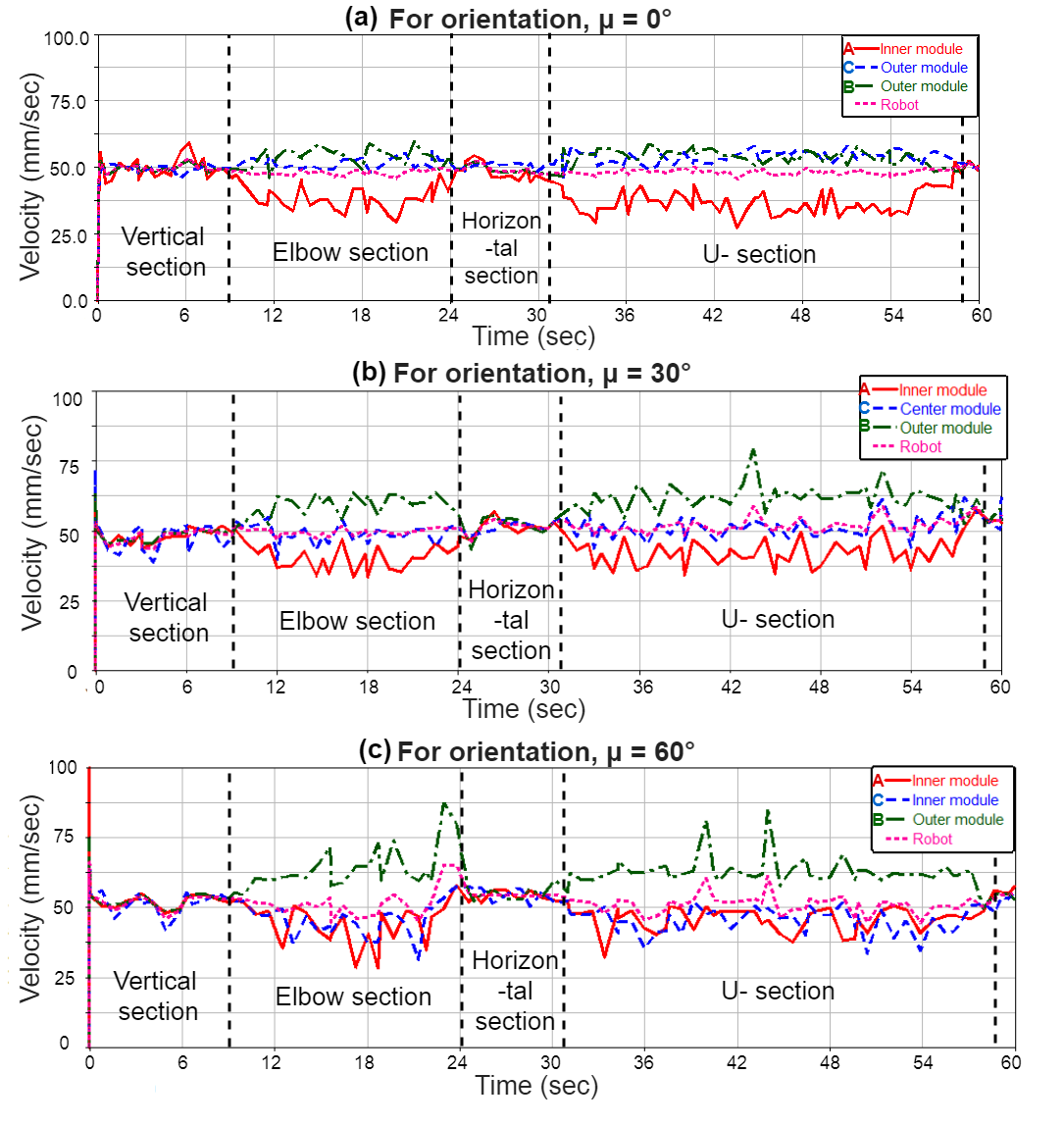}
\vspace{-0.15in}
\caption{Simulation results for (a) $\mu$ = 0$^\circ$, (b) $\mu$ = 30$^\circ$, (c) $\mu$ = 60$^\circ$}
\label{News}
\vspace{-0.2in}
\end{figure}

\vspace{-0.1in}
\subsection{Elbow section and U-section}

In the elbow section ($90^\circ$ bend) and U-section  ($180^\circ$ bend), the robot moves at a constant distance to the center of curvature of the pipe. The 3-OOD mechanism modulates the output speeds of the tracks $v_{tA}$, $v_{tB}$ and $v_{tC}$ according to the distance from the center of curvature of the pipe. In all three orientations of the robot, the outer module tracks rotates faster to travel a longer distance, while the inner module tracks rotates slower to travel the shortest distance than the radius of curvature of the pipe bend. In pipe bends, the simulation velocities of each track is averaged seperately to approximate the observed track velocities without fluctuations. The approximated velocities of each track is then compared with their respective theoretical velocities to find the absolute percentage error (APE), refer \eqref{eqn02}.
  
For the orientation $\mu$ = 0$^\circ$, the outer modules (B and C) move at an average velocity of (58.7 mm/sec and 57.8 mm/sec), while the inner module (A) move at an average velocity of 33.62mm/sec, refer Fig.~\ref{News}(a). These values correspond to the theoretical values $v_{tB}$ = $v_{tC}$ = 58.51 mm/sec and $v_{tA}$ = 33.69mm/sec with an APE of 1.2\%. Similarly, the track velocities $v_{tA}$, $v_{tB}$ and $v_{tC}$ for $\mu$ = 30$^\circ$ presented in Table I, matches with the average value of the simulation results ($v_{tC}$= 50.3 mm/sec, $v_{tB}$= 63.8 mm/sec, $v_{tA}$= 37.3 mm/sec) with an APE of 3.8\%, refer Fig.~\ref{News}(b). Likewise, the track velocity value for $\mu$ = 60$^\circ$, correspond to the simulation results ($v_{tB}$= 68.5 mm/sec, $v_{tA}$= 40.2 mm/sec and $v_{tC}$= 41.3 mm/sec) with an APE of 2.5\%, refer Fig.~\ref{News}(c). In each orientations of both the straight and bend sections, the error value is very minimal and they can be attributed due to external factors in real-world conditions such as friction. Therefore, minimal velocity fluctuations occurs in the simulation plot. From 9 to 24 seconds, the robot negotiates the elbow section (90$^\circ$ bend) of distance 657.83 mm, while it takes 31 to 59 seconds to travel 1315.66 mm in the U-section (180$^\circ$ bend). Figures \ref{Rsim}(c) and \ref{Rsim}(e) shows the robot's capability to traverse in pipe-bends. 

\textbf{\small Eliminating slip and drag:\normalsize}
The simulation results for the track velocities $v_{tA}$, $v_{tB}$, $v_{tC}$ and the robot $v_{R}$ in different orientations ($\mu$ = 0$^\circ$, $\mu$ = 30$^\circ$, $\mu$ = 60$^\circ$), matches with the theoretical results obtained in section III. It is observed from the simulation that in 60 seconds, the robot traverses throughout the pipe network uniformly at the inserted orientation. The result correspond to our theoretical calculation ($D_{R}$/$v$= 3016.49/50.24 = 60.04 sec). This validates that the Three-Output Open Differential eliminates slip and drag in the tracks of the pipe climber in all orientations without any motion losses. In the simulation, the robot is observed without any slip and drag in all orientations, which further impacts in reduced stress influence on the robot and increased motion smoothness.

\vspace{-0.1in}
\begin{figure}[htbp]
\centerline{\includegraphics[scale=0.23]{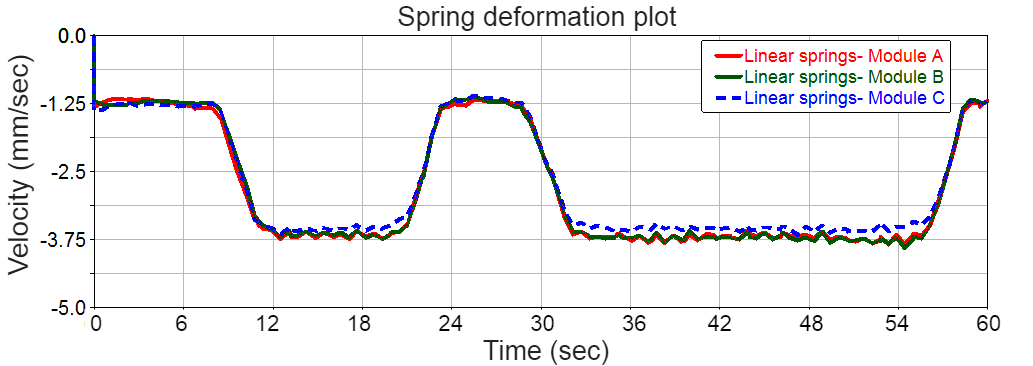}}
\vspace{-0.1in}
\caption{Linear spring deformation in the modules }
\label{Sdef}
\vspace{-0.05in}
\end{figure}

\textbf{\small Radial flexibility:\normalsize}  The track in the modules clamp to the inner wall of the pipe to provide traction during motion. The springs are initially pre-loaded by a compression of 1.25 mm in all three modules equally when inserted in the vertical pipe section, refer Fig. \ref{Sdef}. In straight pipes, the robot moves at the initial pre-loaded spring length. The deformation length increases by 1.5 mm for the inner and the outer modules when the robot is in motion near elbow section and U-section. This deformation explains the radial flexibility allowed for the modules to manoeuvre through the changing cross-section of the pipe diameter in the bends during motion.   

\section{Conclusion and Future works}
The Modular Pipe Climber-III robot is presented with the novel Three-Output Open Differential to control the robot mechanically without any active controls. The differential has an equivalent output to input kinetics, whose performance is completely analogous to the functionality of the traditional two output differential. The simulation results validate successful traversal of complex pipe networks with bends of up to 180$^\circ$ in different orientations without slip. Adopting the differential mechanism in the robot achieves the novel result of eliminating the slip and drag in all orientations of the robot during the motion.

At the present, we are developing a prototype to perform experiments on the proposed design \cite{b26}. In the future, we intend to extend the application of the Three-Output Open Differential for other use cases.


\begin{thebibliography}{00}
\bibitem{b1} Han, Meng, Jun Zhou, Xun Chen, and Lihong Li. "Analysis of in-pipe inspection robot structure design." In 2016 2nd Workshop on Advanced Research and Technology in Industry Applications (WARTIA-16). Atlantis Press, 2016.
\bibitem{b2} Roslin, Nur Shahida, Adzly Anuar, Muhammad Fairuz Abdul Jalal, and Khairul Salleh Mohamed Sahari. "A review: Hybrid locomotion of in-pipe inspection robot." Procedia Engineering 41 (2012): 1456-1462.
\bibitem{b3} Okamoto Jr, Jun, Julio C. Adamowski, Marcos SG Tsuzuki, Flávio Buiochi, and Claudio S. Camerini. "Autonomous system for oil pipelines inspection." Mechatronics 9, no. 7 (1999): 731-743.
\bibitem{b4} Ryew, SungMoo, S. H. Baik, S. W. Ryu, Kwang Mok Jung, S. G. Roh, and Hyouk Ryeol Choi. "In-pipe inspection robot system with active steering mechanism." In Proceedings. 2000 IEEE/RSJ International Conference on Intelligent Robots and Systems (IROS 2000)(Cat. No. 00CH37113), vol. 3, pp. 1652-1657. IEEE, 2000.
\bibitem{b5}Choi, Hyouk Ryeol, and Se-gon Roh. In-pipe robot with active steering capability for moving inside of pipelines. INTECH Open Access Publisher, 2007.

\bibitem{b6}Kakogawa, Atsushi, Taiki Nishimura, and Shugen Ma. "Development of a screw drive in-pipe robot for passing through bent and branch pipes." In IEEE ISR 2013, pp. 1-6. IEEE, 2013.

\bibitem{b10} Hirose, Shigeo, Hidetaka Ohno, Takeo Mitsui, and Kiichi Suyama. "Design of in-pipe inspection vehicles for/spl phi/25,/spl phi/50,/spl phi/150 pipes." In Proceedings 1999 IEEE International Conference on Robotics and Automation (Cat. No. 99CH36288C), vol. 3, pp. 2309-2314. IEEE, 1999.
\bibitem{b12} Debenest, Paulo, Michele Guarnieri, and Shigeo Hirose. "PipeTron series-Robots for pipe inspection." In Proceedings of the 2014 3rd international conference on applied robotics for the power industry, pp. 1-6. IEEE, 2014.
\bibitem{b13} Dertien, Edwin, Mohammad Mozaffari Foumashi, Kees Pulles, and Stefano Stramigioli. "Design of a robot for in-pipe inspection using omnidirectional wheels and active stabilization." In 2014 IEEE International Conference on Robotics and Automation (ICRA), pp. 5121-5126. IEEE, 2014.
\bibitem{b14} Vadapalli, Rama, Kartik Suryavanshi, Ruchita Vucha, Abhishek Sarkar, and K. Madhava Krishna. "Modular Pipe Climber." In Proceedings of the Advances in Robotics 2019, pp. 1-6. 2019.
\bibitem{b15} Suryavanshi, Kartik, Rama Vadapalli, Ruchitha Vucha, Abhishek Sarkar, and K. Madhava Krishna. "Omnidirectional Tractable Three Module Robot." In 2020 IEEE International Conference on Robotics and Automation (ICRA), pp. 9316-9321. IEEE, 2020.
\bibitem{b16} Roh, Se-gon, Jung-Sub Lee, Hyungpil Moon, and Hyouk Ryeol Choi. "Modularized in-pipe robot capable of selective navigation inside of pipelines." In 2008 IEEE/RSJ International Conference on Intelligent Robots and Systems, pp. 1724-1729. IEEE, 2008.
\bibitem{b17} Roh, S. G., SungMoo Ryew, J. H. Yang, and H. R. Choi. "Actively steerable in-pipe inspection robots for underground urban gas pipelines." In Proceedings 2001 ICRA. IEEE International Conference on Robotics and Automation (Cat. No. 01CH37164), vol. 1, pp. 761-766. IEEE, 2001.
\bibitem{b18} Yang, Seung Ung, et al. "Novel robot mechanism capable of 3D differential driving inside pipelines." 2014 IEEE/RSJ International Conference on Intelligent Robots and Systems. IEEE, 2014.
\bibitem{b19} Kim, Ho Moon, Yun Seok Choi, Yoon Geon Lee, and Hyouk Ryeol Choi. "Novel mechanism for in-pipe robot based on a multiaxial differential gear mechanism." IEEE/ASME Transactions on Mechatronics 22, no. 1 (2016): 227-235.
\bibitem{b20} Kim, Ho Moon, Jung Seok Suh, Yun Seok Choi, Tran Duc Trong, Hyungpil Moon, Jachoon Koo, Sungmoo Ryew, and Hyouk Ryeol Choi. "An in-pipe robot with multi-axial differential gear mechanism." In 2013 IEEE/RSJ international conference on intelligent robots and systems, pp. 252-257. IEEE, 2013.
\bibitem{b21} S. Kota and S. Bidare, “Systematic synthesis and applications of novel multi-degree-of-freedom differential systems,” ASME J. Mech, vol. 119, no. 2, pp. 284–291, 1997.
\bibitem{b22} D. Ospina and A. Ramirez-Serrano, “Sensorless in-hand manipulation by an underactuated robot hand,” Journal of Mechanisms and Robotics, vol. 12, no. 5, 2020.
\bibitem{b23} Vadapalli, Rama, Nagamanikandan Govindan, and K. Madhava Krishna. "Design And Analysis Of Three-Output Open Differential with 3-DOF." arXiv preprint arXiv:2112.11254 2021.
\bibitem{b24} J. Deur, V. Ivanovic, M. Hancock, and F. Assadian, “Modeling and´analysis of active differential dynamics,” Journal of dynamic systems, measurement, and control, vol. 132, no. 6, 2010.
\bibitem{b25} J.Scott Armstrong, Fred Collopy. "Error measures for generalizing about forecasting methods: Empirical comparisons." International Journal of Forecasting, Volume 8, Issue 1, 1992, Pages 69-80, ISSN 0169-2070.
\bibitem{b26} Vadapalli, Rama, Saharsh Agarwal, Vishnu Kumar, Kartik Suryavanshi, G. Nagamanikandan, and K. Madhava Krishna. "Modular Pipe Climber III with Three-Output Open Differential." In 2021 IEEE/RSJ International Conference on Intelligent Robots and Systems (IROS), pp. 2473-2478. IEEE, 2021.

\end{thebibliography}
\end{document}